\documentclass[10pt]{article}
\usepackage[margin=0.78in]{geometry}
\usepackage{amsmath,amssymb,amsfonts,mathtools,bm}
\usepackage{graphicx}
\usepackage{xcolor}
\usepackage{enumitem}
\usepackage{booktabs}
\usepackage{natbib}
\usepackage[colorlinks=true,linkcolor=blue,citecolor=blue,urlcolor=blue]{hyperref}
\usepackage{mmll}
\usepackage[bottom]{footmisc}
\usepackage{listings}
\usepackage{url}
\usepackage[table]{xcolor}

\setlist[itemize]{leftmargin=1.4em,itemsep=0.15em,topsep=0.15em}
\setlist[enumerate]{leftmargin=1.4em,itemsep=0.15em,topsep=0.15em}
\definecolor{oursbg}{HTML}{E8F4F5}
\definecolor{ourstext}{HTML}{007681}

\allowdisplaybreaks
\title{Decoupled Marginal Sharpening for Training-Free Inference-Time Scaling}

\author{%
Weixin Wang\textsuperscript{1}\hspace{0.18em}%
\thanks{This work was conducted during Weixin Wang's internship at Morgan Stanley.}%
\enspace \enspace Wei Deng\textsuperscript{2}%
\enspace Anderson Schneider\textsuperscript{2}%
\enspace Yuriy Nevmyvaka\textsuperscript{2}%
\enspace Pan Xu\textsuperscript{1}%
\enspace Andrew Bennett\textsuperscript{2}\\[0.5em]
{\small \textsuperscript{1}Duke University \qquad
\textsuperscript{2}Morgan Stanley}%
}

\begin{document}
\maketitle

\begin{abstract}
We study how to improve LLM reasoning at inference time without updating model weights or using external reward models. Recent reward-free power-sampling methods define the sequence-level power target $\pi_\alpha(y\mid x)\propto p_0(y\mid x)^\alpha$, where $\alpha>1$, which increases the relative probability of completions that are already likely under the base model. Existing methods use Metropolis--Hastings or Sequential Monte Carlo (SMC) to approximately sample from this target. However, two challenges arise at different levels. At the target level, full-sequence sharpening can concentrate probability on a few exact reasoning--answer trajectories, suppressing alternative paths that support the same answer. At the sampler level, finite-particle SMC approximations can suffer from path collapse when resampling repeatedly duplicates a few high-weight prefixes. To address these challenges, we propose \textbf{Decoupled Marginal Sharpening (DMS)}, which separately controls how strongly the model favors particular reasoning paths and how strongly it favors final answers after aggregating support across those paths. At the target level, DMS first samples from a sharpened reasoning distribution, then marginalizes the model's answer probabilities over the resulting weighted reasoning population and sharpens the induced answer marginal. Separate exponents allow mild reasoning sharpening to encourage exploration of diverse paths, followed by stronger answer-marginal sharpening to concentrate probability on answers with greater aggregate support across those paths. At the sampler level, we develop Fully Adapted Annealed SMC, which combines an annealed sequence of power targets with locally fully adapted proposals and predictive ancestor resampling. Experiments on MATH500, GPQA-Diamond, HumanEval, and AIME24/25 across three base models show that DMS achieves the highest mean pass@1 among the compared methods in all tasks.
\end{abstract}

\begin{figure}[h]
    \centering
    \includegraphics[width=\linewidth]{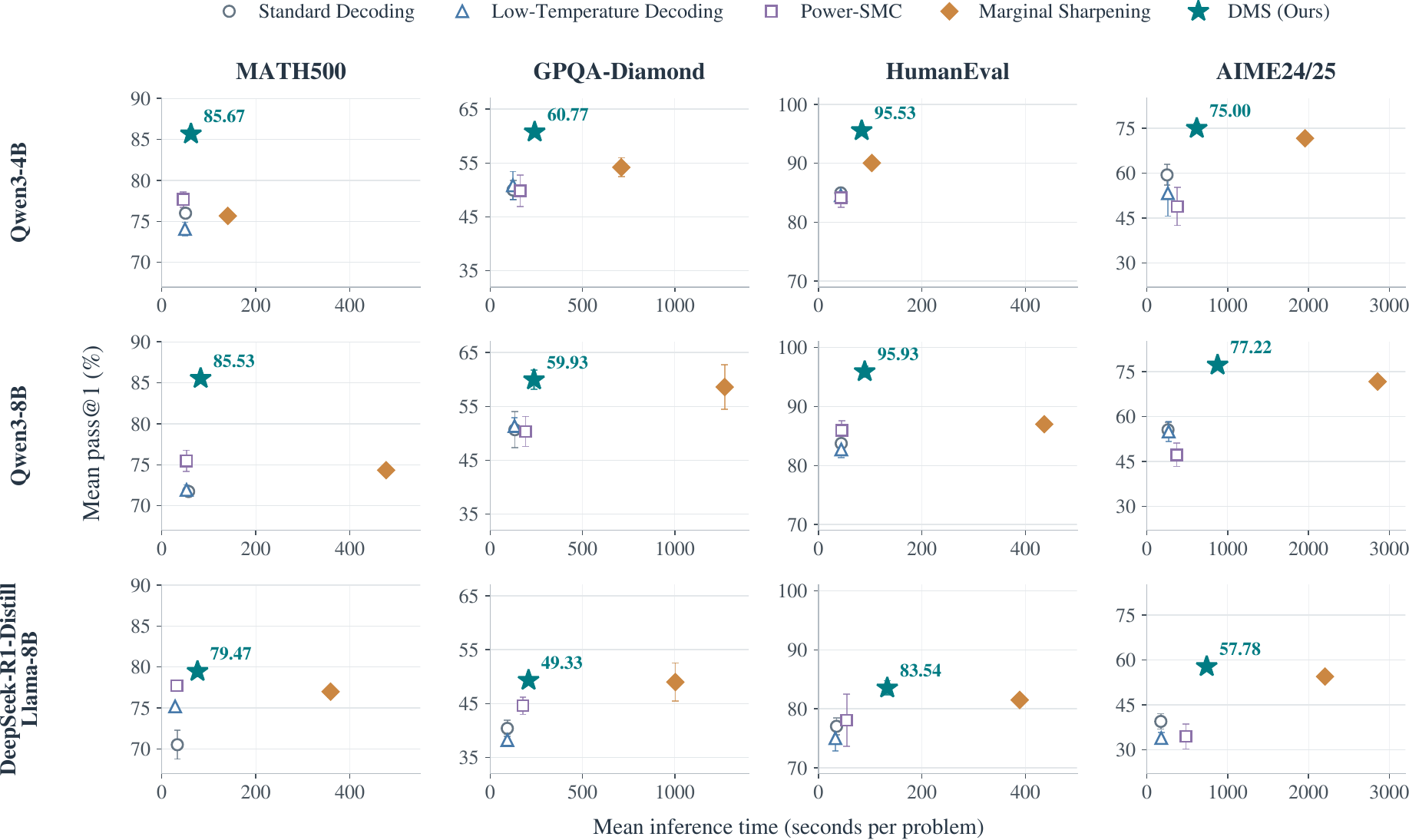}
    \caption{\textbf{Accuracy and inference latency across three models and four reasoning benchmarks.} Our method \textbf{DMS} achieves the highest accuracy in all settings while requiring less inference time than the strong baseline Marginal Sharpening. Vertical error bars indicate $\pm1$ standard deviation in pass@1. In each subplot, higher accuracy and lower latency are better (toward the upper left).}
    \label{fig:accuracy-latency}
\end{figure}

\section{Introduction}

Large language models can improve their reasoning performance through additional training \citep{zelikman2022star,ouyang2022training,hsieh2023distilling,shao2024deepseekmath,zhang2026embarrassingly} or additional computation at inference time \citep{wang2022self,yao2023tree,snell2024scaling,wang2026inference}. We focus on inference-time scaling methods that keep the model fixed and use its own probabilities to guide generation, without additional model updates. 
These methods can be broadly divided into reward-based and reward-free approaches. Reward-based methods use an external reward model or verifier to guide generation toward higher-quality solutions \citep{zhao2024probabilistic,feng2025stepbystep,markovic2026sampling}. Their effectiveness depends on the availability and accuracy of these signals \citep{zhu2026power}. Reward-free methods instead rely on signals already contained in the base model. A prominent approach is sequence-level power sampling, which defines the target distribution $\pi_\alpha(y\mid x) \propto p_0(y\mid x)^\alpha, \alpha>1$, where $p_0$ is the base model, $x$ is the prompt, and $y$ is a complete generated sequence. This power transformation \emph{sharpens} the base distribution by increasing the relative probability of more likely completions, making the distribution more concentrated. Sampling each token according to this target requires summing powered sequence probabilities over all possible future continuations, which is generally intractable. Existing methods approximate this sharpened distribution through Metropolis--Hastings updates to generated sequences \citep{karan2025reasoning}, token sampling guided by probabilities estimated from future continuations \citep{ji2026scalable}, or Sequential Monte Carlo (SMC) \citep{azizi2026power}.

For mathematical and coding tasks, an LLM typically generates a reasoning path followed by a final answer. Sharpening the full sequence therefore favors individual reasoning--answer pairs with high joint likelihood. However, different reasoning paths can lead to the same correct answer, which may receive substantial combined support even when each individual path has relatively low probability. Sharpening these pairs separately can consequently reduce the probability of an answer supported by many such paths. Moreover, higher sequence likelihood does not necessarily imply correctness \citep{zenn2026likely}. Self-consistency and marginal sharpening account for support across reasoning paths \citep{wang2022self,arzhantsev2026self}; in particular, marginal sharpening aggregates conditional answer probabilities under the original reasoning distribution before sharpening the resulting answer marginal. We seek to retain this aggregation while also guiding reasoning toward more likely paths. The two stages thus call for different sharpening strengths: mild sharpening during reasoning favors more likely paths while retaining alternatives for aggregation, whereas stronger sharpening after aggregation concentrates probability on answers supported by their combined contributions.

To this end, we propose \textbf{Decoupled Marginal Sharpening (DMS)}, a training-free framework that separately controls sharpening at the reasoning and answer stages. Given a prompt $x$, let $z$ denote a reasoning path and $a$ denote its answer. DMS first sharpens the reasoning distribution, averages the model's conditional answer probabilities over the resulting distribution of reasoning paths, and then sharpens the resulting answer marginal:
\begin{align*}
    \textstyle \pi_Z(z| x) \propto p_0(z| x)^{\alpha_Z}, \quad m_{\alpha_Z}(a| x) = \sum_z \pi_Z(z| x)p_0(a| z,x), \quad \pi_A(a| x) \propto m_{\alpha_Z}(a| x)^{\alpha_A}.
\end{align*}
The reasoning exponent $\alpha_Z$ controls how much each reasoning path contributes to the answer marginal, while the answer exponent $\alpha_A$ controls how strongly the final distribution concentrates on answers with greater aggregate support. Choosing these exponents separately allows answer selection to become more concentrated without requiring equally strong sharpening of the reasoning distribution. In practice, we estimate the answer marginal from a finite population of reasoning paths. Repeated SMC resampling can duplicate a few high-weight prefixes and discard alternatives, causing path collapse and reducing the distinct reasoning paths available for aggregation. To address this sampling challenge, we develop Fully Adapted Annealed SMC (FAA-SMC), which combines annealed power targets, predictive ancestor resampling, and locally fully adapted proposals. The resulting weighted reasoning population defines an empirical answer marginal. DMS samples answer candidates from this marginal, evaluates each complete candidate under all reasoning paths, and approximates the sharpened answer target.

\textbf{Our contributions} are summarized as follows:
\begin{itemize}[leftmargin=10pt,nosep]
    \item We propose \textbf{DMS}, a training-free two-stage marginal sharpening framework with separate control over reasoning-path concentration and answer-level consensus. DMS sharpens the reasoning distribution and then sharpens the answer marginal obtained by aggregating probabilities across reasoning paths, using different exponents for the two stages.

    \item We develop a practical sampling procedure for the two-stage target. FAA-SMC combines exact annealing bridges, predictive resampling, and locally fully adapted propagation to approximate the sharpened reasoning distribution. A finite-particle answer sampler then approximates the sharpened answer marginal through cross-reasoning likelihood evaluation.

    \item We evaluate DMS on MATH500, GPQA-Diamond, HumanEval, and AIME24/25 across three base models. DMS achieves the highest mean pass@1 among the compared methods in all twelve model--benchmark settings.
\end{itemize}

\section{Related Work}

\paragraph{Inference-Time Scaling for LLM Reasoning.}

Inference-time scaling improves LLM reasoning by allocating additional computation during decoding while keeping the base model fixed. Existing methods can be broadly divided into reward-based and reward-free approaches. Reward-based methods use an external reward or verifier to define a reward-tilted sequence distribution. Building on auxiliary and twisted particle filtering \citep{guarniero2017iterated}, \citet{zhao2024probabilistic} formulate language-model inference as sampling from unnormalized sequence distributions and introduce twisted SMC with learned estimates of future potential. \citet{feng2025stepbystep} apply this framework to mathematical reasoning, while \citet{markovic2026sampling} develop a general reward-guided SMC decoder with prefix-based and look-ahead intermediate targets. \citet{zhu2026power} further study how errors in approximate reward models affect inference-time scaling.

Reward-free methods instead use signals already contained in the base model, such as sequence likelihood or agreement among sampled solutions. One line of work uses sequence-level power sampling, which defines the target $\pi_\alpha(y\mid x)\propto p_0(y\mid x)^\alpha$ for $\alpha>1$. \citet{karan2025reasoning} approximately sample from this target using an iterative Metropolis--Hastings-style procedure. \citet{ji2026scalable} improve sampling efficiency by approximating the future correction associated with the sequence-level power objective during autoregressive decoding. Sequential Monte Carlo provides a parallel alternative by maintaining and resampling a population of weighted partial generations. \citet{azizi2026power} propose Power-SMC, which uses token-level importance corrections, adaptive resampling, and exponent bridging to approximate the sequence-level power target. \citet{nguyen2026model} further improve particle allocation using short-horizon rollouts or an amortized estimate of future probability mass. Related reward-free SMC methods also consider sampling from ensemble distributions constructed from multiple language models or prompts \citep{chan2026ensembling}. Complementary studies examine how model-intrinsic scores interact with sampling, optimization, and consensus-based decoders \citep{cheng2026score}, and show that increasing sequence likelihood does not necessarily improve reasoning accuracy \citep{zenn2026likely}.

Beyond sampling individual sequences, reward-free methods can aggregate support across reasoning paths at the answer level. Self-consistency samples multiple reasoning trajectories and selects the answer with the greatest aggregate support \citep{wang2022self}. \citet{arzhantsev2026self} formulate self-consistency as marginal sharpening over final answers, constructing the answer marginal under the original reasoning distribution. DMS extends this formulation by allowing the reasoning distribution used to construct the marginal to be sharpened independently of the subsequent answer sharpening.

\section{Preliminaries}
\label{sec:preliminary}

\paragraph{LLM Generation and Sequence-Level Power Sharpening.}

Given a problem prompt $x$, an autoregressive language model $p_0$ generates each token conditioned on the prompt and the preceding tokens. The probability of a sequence $y_{1:T}=(y_1,\dots,y_T)$ is therefore $p_0(y_{1:T}\mid x)=\prod_{t=1}^T p_0(y_t\mid x,y_{<t})$, where $y_{<t}=(y_1,\dots,y_{t-1})$. We use $T$ as the maximum generation length. If generation ends before $T$, we pad the sequence with a fixed terminal symbol and assign conditional probability one to each padding step, leaving the sequence probability unchanged.

\citet{huang2025self} observe that correct responses tend to have higher likelihoods under the model and that selecting responses with higher likelihoods can improve accuracy, although the gains depend on the base model and task. These observations motivate \emph{distribution sharpening}: keeping the model parameters fixed while changing the sampling distribution to favor responses that the model already assigns higher probability.

For training-free LLM reasoning, \citet{karan2025reasoning} propose sampling from a sequence-level power distribution that implements this principle while keeping the base model fixed:
\begin{align}
    \label{equ:power_target}
    \pi_\alpha(y_{1:T}\mid x) = \mathcal{Z}_\alpha(x)^{-1}p_0(y_{1:T}\mid x)^\alpha, \qquad \alpha\geq 1,
\end{align}
where $\mathcal{Z}_\alpha(x)=\sum_{y_{1:T}}p_0(y_{1:T}\mid x)^\alpha$ is the normalizing constant. When $\alpha=1$, this target recovers the base model distribution. For $\alpha>1$, it favors outputs the model is more confident in, as reflected by their higher likelihood.

To sample a sequence from $\pi_\alpha$ one token at a time, we need the conditional probability of each next token under $\pi_\alpha$. Computing these probabilities requires summing $p_0(y_{1:T}\mid x)^\alpha$ over all possible future continuations, which is generally intractable. Simply lowering the temperature of the base model's next-token distribution does not account for the powered probability mass of these continuations and therefore does not generally produce samples from $\pi_\alpha$. We therefore use sequential Monte Carlo (SMC), which generates tokens from tractable proposal distributions and uses importance weights to correct for the difference between the proposal and target distributions.

\paragraph{Sequential Monte Carlo.}
SMC approximates the target distribution using $K$ particles, each consisting of a generated prefix and an importance weight \citep{zhao2024probabilistic,azizi2026power}. At step $t$, the weighted particles approximate the distribution proportional to $\gamma_t(y_{1:t}\mid x)$, where $\{\gamma_t\}_{t=0}^T$ is a sequence of unnormalized intermediate targets satisfying
\begin{align*}
    \gamma_0(\varnothing\mid x)=1, \qquad \gamma_T(y_{1:T}\mid x)=p_0(y_{1:T}\mid x)^\alpha.
\end{align*}
The normalized terminal target is therefore $\pi_\alpha$ in \eqref{equ:power_target}.

All particles start from the empty prefix with equal weights $w_0^k=1$. At each step, a proposal distribution $q_t(\cdot\mid x,y_{<t})$ specifies how to sample the next token. To account for the proposal probability and the change from $\gamma_{t-1}$ to $\gamma_t$, we use the incremental importance weight
\begin{align}
    \label{equ:smc_incremental_weight}
    G_t(y_{1:t}) = \frac{\gamma_t(y_{1:t}\mid x)}{\gamma_{t-1}(y_{<t}\mid x)q_t(y_t\mid x,y_{<t})}.
\end{align}
Particle $k$ is then extended and reweighted according to
\begin{align*}
    Y_t^k \sim q_t(\cdot\mid x,Y_{<t}^k), \qquad w_t^k = w_{t-1}^kG_t(Y_{1:t}^k).
\end{align*}

To determine when to resample, we normalize the weights and compute the effective sample size:
\begin{align*}
    \bar w_t^k = \frac{w_t^k}{\sum_{\ell=1}^K w_t^\ell}, \qquad \operatorname{ESS}_t = \frac{1}{\sum_{k=1}^K(\bar w_t^k)^2}.
\end{align*}
A small ESS indicates that a few particles carry most of the total weight. When $\operatorname{ESS}_t<\tau K$ for a threshold $\tau\in(0,1]$, SMC selects $K$ prefixes according to their normalized weights, allowing the same prefix to be selected multiple times. The selected prefixes form a new particle population with equal weights, and generation continues separately from each selected prefix. At termination, the weighted population $\{(Y_{1:T}^k,\bar w_T^k)\}_{k=1}^K$ approximates $\pi_\alpha$.

With a finite particle budget, repeated resampling can lead to \emph{path collapse}: high-weight prefixes are repeatedly copied while other prefixes are discarded, leaving many particles descended from only a few earlier prefixes. Although resampling resets the weights, it does not recover the discarded paths. This loss of reasoning-path diversity motivates the sampler design introduced in our methodology.

\section{Methodology: Decoupled Marginal Sharpening}
\label{sec:methodology_tms}

\begin{figure}[t]
    \centering
    \includegraphics[width=\linewidth]{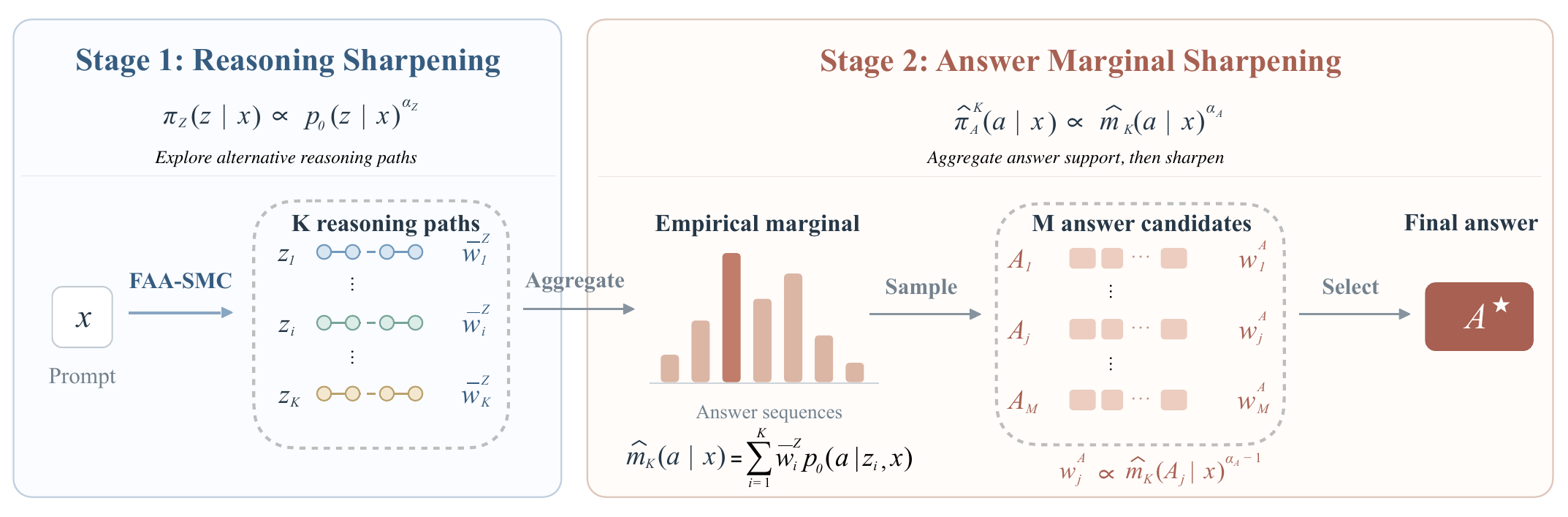}
    \caption{Overview of DMS. FAA-SMC constructs $K$ weighted reasoning paths that define an empirical answer marginal. We sample $M$ candidates from this marginal and select the final answer according to their normalized answer-sharpening weights.}
    \label{fig:dms_flow_chart}
\end{figure}

We propose \textbf{Decoupled Marginal Sharpening (DMS)}, which first sharpens the reasoning distribution and then sharpens the answer marginal constructed from it. To approximate these targets, we develop Fully Adapted Annealed SMC (FAA-SMC) to generate a weighted population of reasoning paths. We sample answer candidates from the resulting empirical marginal, evaluate each candidate under all paths, and reweight them for final selection. \Cref{fig:dms_flow_chart} illustrates the overall procedure.

\subsection{Two-Stage Sharpened Target}
\label{sec:tms_target}

We write the complete generation as $Y=(Z,A)$, where $Z$ is the reasoning path and $A$ is the complete answer token sequence generated after it, such as a mathematical response or a code completion. The base model distribution factorizes as $p_0(z,a\mid x)=p_0(z\mid x)p_0(a\mid z,x)$.

Sequence-level power sharpening applies the same exponent to both factors: $p_0(z,a\mid x)^\alpha=p_0(z\mid x)^\alpha p_0(a\mid z,x)^\alpha$. This ties reasoning and conditional answer sharpening to the same strength. However, reasoning and answer selection serve different purposes. During reasoning, we want to favor likely paths while retaining alternatives for exploration; when selecting the final answer, we want to concentrate on candidates with strong model support. We therefore introduce separate exponents, allowing mild or no additional reasoning sharpening together with stronger answer sharpening.

\paragraph{Reasoning-Level Sharpening.}
We first define the reasoning target with its own exponent:
\begin{align}
    \label{equ:tms_reasoning_target}
    \pi_Z(z\mid x) = \mathcal{Z}_Z(x)^{-1}p_0(z\mid x)^{\alpha_Z}, \qquad \alpha_Z\geq 1,
\end{align}
where $\mathcal{Z}_Z(x)=\sum_z p_0(z\mid x)^{\alpha_Z}$ is the normalizing constant. Setting $\alpha_Z=1$ retains the base reasoning distribution, while increasing $\alpha_Z$ gives greater relative probability to more likely reasoning paths. Choosing this exponent independently lets us control the preference for these paths during reasoning exploration without tying it to the strength used for answer selection.

\paragraph{Reasoning-Induced Answer Marginal.}
With the reasoning distribution in \eqref{equ:tms_reasoning_target} specified, we next construct the corresponding answer marginal. Marginalization combines support for the same answer across different reasoning paths, so that subsequent answer sharpening acts on their total contribution. We define the answer marginal distribution as follows,
\begin{align*}
\textstyle
    m_{\alpha_Z}(a\mid x) = \sum_z \pi_Z(z\mid x)p_0(a\mid z,x).
\end{align*}
The marginal measures the combined support for this answer under chosen reasoning distribution.

\paragraph{Marginal Answer Sharpening.}
We then sharpen the answer marginal with a separate exponent:
\begin{align}
    \label{equ:tms_answer_target}
    \pi_A(a\mid x) = \mathcal{Z}_A(x)^{-1}m_{\alpha_Z}(a\mid x)^{\alpha_A}, \qquad \alpha_A\geq 1,
\end{align}
where $\mathcal{Z}_A(x)=\sum_a m_{\alpha_Z}(a\mid x)^{\alpha_A}$ is the normalizing constant. For a fixed $\alpha_Z$, increasing $\alpha_A$ gives greater relative probability to answers with larger marginal probability while preserving their ranking under $m_{\alpha_Z}$. The final answer distribution can therefore be sharpened more strongly while the reasoning distribution remains fixed.

\begin{remark}
The order of marginalization and answer sharpening matters. Under sequence-level sharpening, the probability assigned to an answer is proportional to $\sum_z p_0(z,a\mid x)^\alpha$, so each trajectory is sharpened before contributions to the same answer are combined. This can make a single likely trajectory outweigh several less likely trajectories even when their combined probability was originally greater, reversing the ranking of their associated answers.\footnote{Consider a simple example: suppose three trajectories leading to answer $a$ each have probability $0.2$, while one trajectory leading to answer $b$ has probability $0.4$. Their base answer probabilities are $0.6$ and $0.4$. With $\alpha=2$, sequence-level sharpening gives unnormalized answer masses $0.12$ and $0.16$, reversing their ranking. Sharpening the base answer marginal instead gives $0.36$ and $0.16$.} In DMS, answer sharpening acts on the combined probability already obtained under $\pi_Z$. Consequently, even setting $\alpha_Z=\alpha_A>1$ does not generally recover the answer marginal of sequence-level sharpening. When $\alpha_Z=1$, we have $m_1(a\mid x)=p_0(a\mid x)$, and the ideal target in \eqref{equ:tms_answer_target} reduces to marginal sharpening under the original reasoning distribution \citep{arzhantsev2026self}. 
\end{remark}

To approximate the target in \eqref{equ:tms_answer_target}, we first estimate $m_{\alpha_Z}$ using a finite weighted population of reasoning paths. We next describe FAA-SMC, a stable and efficient sampler designed to mitigate path collapse when constructing this population.

\subsection{Reasoning Sampling with FAA-SMC}
\label{sec:methodology_faa_smc}

FAA-SMC approximates the reasoning target in \eqref{equ:tms_reasoning_target} with final exponent $\alpha=\alpha_Z$. It combines two mechanisms: annealing introduces sharpening gradually during generation, while a resample--propagate update allows copies of a selected prefix to generate their next tokens independently.

\paragraph{Annealed Intermediate Targets.}
Applying the final sharpening strength from the start can concentrate particles before alternative reasoning paths have been explored. We therefore use a nondecreasing schedule $1=\beta_0\leq\beta_1\leq\cdots\leq\beta_T=\alpha$ and define
\begin{align}
    \label{equ:faa_intermediate_target}
    \gamma_t(y_{1:t}\mid x) = p_0(y_{1:t}\mid x)^{\beta_t}.
\end{align}
Under these targets, the incremental importance weight in \eqref{equ:smc_incremental_weight} becomes
\begin{align}
    \label{equ:faa_incremental_weight}
    G_t(y_{1:t}) = p_0(y_{<t}\mid x)^{\beta_t-\beta_{t-1}}\frac{p_0(y_t\mid x,y_{<t})^{\beta_t}}{q_t(y_t\mid x,y_{<t})}.
\end{align}
The prefix factor accounts for the change in exponent applied to tokens already generated. Since $\beta_T=\alpha$, \eqref{equ:faa_intermediate_target} ends at the specified power target. Annealing therefore changes how sharpening is introduced during generation while preserving the final target.

\paragraph{Fully Adapted Resample--Propagate.}
We next choose a proposal that allows the importance correction to be computed before generating the next token. Let $h_t=(x,y_{<t})$ denote the current context. We use the locally fully adapted proposal
\begin{align*}
    q_t^{\mathrm{FA}}(v\mid h_t) = \frac{p_0(v\mid h_t)^{\beta_t}}{\mathcal{Z}_t(h_t)},
\end{align*}
where $\mathcal{Z}_t(h_t)=\sum_{v\in\mathcal{V}}p_0(v\mid h_t)^{\beta_t}$ and $\mathcal{V}$ is the token vocabulary. Substituting this proposal into \eqref{equ:faa_incremental_weight} cancels the dependence on the sampled token:
\begin{align*}
    G_t(y_{1:t}) = M_t(h_t), \qquad M_t(h_t) = p_0(y_{<t}\mid x)^{\beta_t-\beta_{t-1}}\mathcal{Z}_t(h_t).
\end{align*}
The remaining correction depends only on the current prefix. We can therefore compute the predictive weights $\widetilde w_{t-1}^k=w_{t-1}^kM_t(h_t^k)$ and use them to select ancestor prefixes before propagation.

At resampling checkpoints, we resample prefixes according to their normalized predictive weights if $\operatorname{ESS}<\tau K$ and reset their weights to equal values. Each prefix then independently samples its next token from $q_t^{\mathrm{FA}}$, with no further weight correction since $M_t$ already accounts for it. Resampling before propagation allows copies of the same prefix to generate different next tokens instead of inheriting the same newly sampled token. Algorithm~\ref{alg:faa_smc} in Appendix~\ref{app:dms-details} summarizes the procedure.

\begin{remark}[Comparison with propagate--resample SMC]
Consider a prefix selected twice at a resampling step. A propagate--resample update in standard SMC copies the same newly generated token into both particles. FAA-SMC instead selects the prefix first and then draws a next token independently for each copy, allowing the copies to follow different continuations immediately. For example, if two tokens each have probability $1/2$ under $q_t^{\mathrm{FA}}$, two copies of the same prefix independently draw different tokens with probability $1/2$.
\end{remark}

\paragraph{Island-DMS.}
On difficult problems requiring long reasoning paths, repeated resampling can lead to more severe path collapse, especially when the particle population is small. In extreme cases, all final particles may descend from only one or two initial particles. To preserve multiple particle diversity, we adopt an island variant \citep{verge2015parallel}. We divide the total population into $L$ independent islands of $K_{\mathrm{island}}$ particles, with $K=L K_{\mathrm{island}}$. Each island runs FAA-SMC and resamples only within its own population. Since each island retains at least one initial-particle ancestor, pooling all islands guarantees at least $L$ distinct ancestors, even if all particles within each island share the same ancestor.

We retain all final reasoning paths and assign equal total weight to each island. If $\bar w_{\ell,k}^Z$ is the normalized weight of particle $k$ within island $\ell$, its pooled weight is $\bar w_{\ell,k}^{Z,\mathrm{island}}=\bar w_{\ell,k}^Z/L$. This pooled population defines the empirical answer marginal used in the next subsection.

\subsection{Sampling from the Sharpened Answer Marginal}
\label{sec:tms_answer_sampling}

Let $\mathcal{B}_K^Z=\{(z_i,\bar w_i^Z)\}_{i=1}^K$ denote the weighted reasoning population returned by FAA-SMC, with $\sum_{i=1}^K\bar w_i^Z=1$. For Island-DMS, we use the pooled population defined above. Each path ends at the reasoning--answer boundary and provides the context for both answer generation and scoring. We replace the sum over all reasoning paths in $m_{\alpha_Z}$ with the weighted sum over this population:
\begin{align}
    \label{equ:tms_empirical_answer_marginal}
    \textstyle
    \widehat m_K(a\mid x) = \sum_{i=1}^K \bar w_i^Z p_0(a\mid z_i,x).
\end{align}
This empirical marginal defines the answer target $\widehat\pi_A^K(a\mid x)\propto\widehat m_K(a\mid x)^{\alpha_A}$, which approximates the ideal target in \eqref{equ:tms_answer_target}. We approximate sampling from this target by generating candidates from $\widehat m_K$ and correcting the difference through importance weighting.

\paragraph{Generating Answer Candidates.}
Since $\widehat m_K$ is a mixture, we can sample from it by first selecting a reasoning path according to its weight and then generating an answer conditioned on that path. Repeating these two draws independently produces $M$ candidates:
\begin{align*}
    I_j \sim \operatorname{Cat}(\bar w_{1:K}^Z), \qquad A_j \sim p_0(\cdot\mid z_{I_j},x), \qquad j=1,\dots,M.
\end{align*}
Each candidate therefore follows $\widehat m_K$, even though its generation uses only one reasoning path. This mixture serves as the proposal distribution for answer sharpening.

\paragraph{Evaluating Answer Marginal Probabilities.}
Generating a candidate requires only one selected path, but evaluating its marginal probability requires contributions from the entire reasoning population. We therefore score the same complete answer $A_j$ under every retained context $(x,z_i)$ and combine the conditional probabilities $p_0(A_j\mid z_i,x)$ using \eqref{equ:tms_empirical_answer_marginal}. This allows paths other than the one used for generation to contribute to the candidate's probability for final selection. Additional detail of derivation is described in Appendix~\ref{app:dms-details}.

\paragraph{Answer Reweighting and Selection.}
The proposal is $\widehat m_K$, whereas the unnormalized answer target is $\widehat m_K^{\alpha_A}$. Their ratio gives the importance weight
\begin{align}
    \label{equ:tms_answer_importance_weight}
    w_j^A = \frac{\widehat m_K(A_j\mid x)^{\alpha_A}}{\widehat m_K(A_j\mid x)} = \widehat m_K(A_j\mid x)^{\alpha_A-1}.
\end{align}
Candidate generation already samples answers in proportion to their marginal probabilities, so the correction uses only the remaining power $\alpha_A-1$. We normalize the weights as $\bar w_j^A=w_j^A/\sum_{\ell=1}^M w_\ell^A$, draw $J^\star\sim\operatorname{Cat}(\bar w_{1:M}^A)$, and return $A^\star=A_{J^\star}$.

Conditional on the reasoning population, this sampling-importance-resampling procedure approximates $\widehat\pi_A^K$ using $M$ candidates. Algorithm~\ref{alg:fp_mas} summarizes DMS.

\begin{algorithm}[t]
\caption{DMS: Decoupled Marginal Sharpening}
\label{alg:fp_mas}
\begin{algorithmic}[1]
\REQUIRE Prompt $x$, base model $p_0$, reasoning particle count $K$, answer candidate count $M$, exponents $\alpha_Z,\alpha_A$, reasoning schedule $1=\beta_0\leq\cdots\leq\beta_T=\alpha_Z$, resampling threshold $\tau$, and resampling checkpoints.
\STATE Run FAA-SMC in Algorithm~\ref{alg:faa_smc} over the reasoning segment to obtain $\mathcal{B}_K^Z=\{(z_i,\bar w_i^Z)\}_{i=1}^K$.
\FOR{$j=1,\dots,M$}
    \STATE Sample $I_j\sim\operatorname{Cat}(\bar w_{1:K}^Z)$.
    \STATE Generate a complete answer $A_j\sim p_0(\cdot\mid z_{I_j},x)$.
    \STATE Score $A_j$ under all $K$ reasoning paths and compute $\widehat m_K(A_j\mid x)$ using \eqref{equ:tms_empirical_answer_marginal}.
    \STATE Compute $w_j^A$ using \eqref{equ:tms_answer_importance_weight}.
\ENDFOR
\STATE Normalize the answer weights to obtain $\bar w_{1:M}^A$.
\STATE Sample $J^\star\sim\operatorname{Cat}(\bar w_{1:M}^A)$ and \textbf{return} $A^\star=A_{J^\star}$.
\end{algorithmic}
\end{algorithm}

\section{Experiments}

\label{sec:experiments}

\subsection{Experimental Setup}

\paragraph{Models and baselines.}
We evaluate DMS using Qwen3-4B, Qwen3-8B \citep{yang2025qwen3}, and DeepSeek-R1-Distill-Llama-8B \citep{guo2025deepseek}. The Qwen3 models use their native chat templates with thinking mode enabled. We compare against standard decoding, low-temperature decoding, Power-SMC \citep{azizi2026power}, and Marginal Sharpening \citep{arzhantsev2026self}. Power-SMC targets a sharpened distribution over complete sequences, while Marginal Sharpening sharpens the answer marginal induced by the original reasoning distribution.

\paragraph{Benchmarks and evaluation.}
We evaluate on MATH500 (500 problems) \citep{hendrycks2021measuring,lightman2024let}, GPQA-Diamond (198 questions) \citep{rein2023gpqa}, HumanEval (164 problems) \citep{chen2021evaluating}, and AIME24/25 (60 problems from AIME I and II in both years). These benchmarks cover mathematical reasoning, scientific question answering, and code generation. Mathematical predictions are extracted from the final boxed expression and graded using normalized exact matching and symbolic equivalence. GPQA-Diamond uses exact option matching, while HumanEval completions are evaluated against the original unit tests in isolated processes with a three-second timeout.

\paragraph{Sampling protocol.}
DMS uses $K=16$ reasoning particles and $M=32$ answer candidates per problem. We use standard DMS on MATH500, GPQA-Diamond, and HumanEval, and Island-DMS with eight islands of two particles on AIME24/25. Candidates are sampled from the weighted reasoning mixture and cross-scored under all $K$ paths to estimate their marginal probabilities and compute answer-level importance weights. Generation budgets, sharpening exponents, and implementation details are provided in Appendix~\ref{app:experimental-details}.

\paragraph{Metrics.}
Our primary metric is pass@1 of the final returned answer. For DMS, one answer is sampled categorically from the $M$ candidates according to their normalized importance weights. We report the mean and sample standard deviation over three random seeds, $\{10080,10081,10082\}$.

\subsection{Main Results}

\begin{table*}[t]
\centering
\small
\setlength{\tabcolsep}{7pt}
\renewcommand{\arraystretch}{1.12}
\caption{Pass@1 (\%) across models and benchmarks. Shaded rows denote DMS, and boldface indicates the highest mean within each model--benchmark setting. The DMS results on
AIME24/25 use the Island-DMS variant.}
\label{tab:main-results}
\resizebox{\textwidth}{!}{
\begin{tabular}{llcccc}
\toprule
\textbf{Model} & \textbf{Method} & \textbf{MATH500}
& \textbf{GPQA-Diamond} & \textbf{HumanEval} & \textbf{AIME24/25} \\
\midrule

\multirow{5}{*}{\textbf{Qwen3-4B}}
& Standard Decoding
    & \(76.00 \pm 0.53\)
    & \(50.00 \pm 1.75\)
    & \(84.96 \pm 0.35\)
    & \(59.44 \pm 3.47\) \\
& Low-Temperature Decoding
    & \(74.07 \pm 0.81\)
    & \(50.84 \pm 2.59\)
    & \(84.55 \pm 1.27\)
    & \(53.33 \pm 7.64\) \\
& Power-SMC
    & \(77.67 \pm 0.95\)
    & \(49.83 \pm 2.92\)
    & \(84.15 \pm 1.61\)
    & \(48.89 \pm 6.31\) \\
& Marginal Sharpening
    & \(75.67 \pm 0.23\)
    & \(54.21 \pm 1.77\)
    & \(90.04 \pm 0.93\)
    & \(71.67 \pm 1.67\) \\
\rowcolor{oursbg}
& \textcolor{ourstext}{\textbf{DMS (Ours)}}
    & \(\mathbf{85.67 \pm 0.58}\)
    & \(\mathbf{60.77 \pm 0.77}\)
    & \(\mathbf{95.53 \pm 0.70}\)
    & \(\mathbf{75.00 \pm 1.67}\) \\
\midrule

\multirow{5}{*}{\textbf{Qwen3-8B}}
& Standard Decoding
    & \(71.73 \pm 0.58\)
    & \(50.67 \pm 3.36\)
    & \(83.74 \pm 0.35\)
    & \(55.56 \pm 2.55\) \\
& Low-Temperature Decoding
    & \(71.93 \pm 0.61\)
    & \(51.35 \pm 1.54\)
    & \(82.72 \pm 1.41\)
    & \(55.00 \pm 3.33\) \\
& Power-SMC
    & \(75.47 \pm 1.29\)
    & \(50.34 \pm 2.78\)
    & \(85.98 \pm 1.61\)
    & \(47.22 \pm 3.85\) \\
& Marginal Sharpening
    & \(74.33 \pm 0.12\)
    & \(58.59 \pm 4.13\)
    & \(86.99 \pm 0.70\)
    & \(71.67 \pm 1.67\) \\
\rowcolor{oursbg}
& \textcolor{ourstext}{\textbf{DMS (Ours)}}
    & \(\mathbf{85.53 \pm 0.42}\)
    & \(\mathbf{59.93 \pm 1.77}\)
    & \(\mathbf{95.93 \pm 0.93}\)
    & \(\mathbf{77.22 \pm 0.96}\) \\
\midrule

\multirow{5}{*}{\textbf{DeepSeek-R1-Distill-Llama-8B}}
& Standard Decoding
    & \(70.53 \pm 1.75\)
    & \(40.40 \pm 1.52\)
    & \(77.03 \pm 1.41\)
    & \(39.44 \pm 2.55\) \\
& Low-Temperature Decoding
    & \(75.20 \pm 0.20\)
    & \(38.22 \pm 1.05\)
    & \(75.00 \pm 2.11\)
    & \(33.89 \pm 1.92\) \\
& Power-SMC
    & \(77.73 \pm 0.42\)
    & \(44.61 \pm 1.62\)
    & \(78.05 \pm 4.40\)
    & \(34.44 \pm 4.19\) \\
& Marginal Sharpening
    & \(77.00 \pm 0.69\)
    & \(48.99 \pm 3.54\)
    & \(81.50 \pm 0.93\)
    & \(54.44 \pm 1.92\) \\
\rowcolor{oursbg}
& \textcolor{ourstext}{\textbf{DMS (Ours)}}
    & \(\mathbf{79.47 \pm 0.61}\)
    & \(\mathbf{49.33 \pm 0.77}\)
    & \(\mathbf{83.54 \pm 1.22}\)
    & \(\mathbf{57.78 \pm 0.96}\) \\

\bottomrule
\end{tabular}
}
\end{table*}

Table~\ref{tab:main-results} reports pass@1 across three models and four benchmarks. DMS achieves the highest mean accuracy in all twelve settings, improving over the strongest baseline by 0.34--10.07 percentage points. On MATH500, DMS reaches 85.67\% and 85.53\% with Qwen3-4B and Qwen3-8B, exceeding Power-SMC by 8.00 and 10.07 points, respectively. On HumanEval, it achieves 95.53\% and 95.93\%, improving over Marginal Sharpening by 5.49 and 8.94 points.

Standard DMS achieves the highest mean accuracy on MATH500, GPQA-Diamond, and HumanEval for all three models. It also exceeds the strongest baseline on AIME24/25, reaching 72.78\%, 74.44\%, and 55.00\% with Qwen3-4B, Qwen3-8B, and DeepSeek-R1-Distill-Llama-8B, respectively. Island-DMS further raises these results to 75.00\%, 77.22\%, and 57.78\%, adding 2.22--2.78 percentage points while retaining the same total reasoning particle count.

The strongest baseline varies across tasks: Power-SMC performs best on MATH500, while Marginal Sharpening leads on the other three benchmarks. DMS achieves the highest mean pass@1 in each setting. Its comparison with Marginal Sharpening is particularly relevant because this baseline already aggregates answer probabilities across reasoning paths. The results support the complete DMS procedure, which combines a separately controlled reasoning distribution with marginal-probability weighting of complete answer candidates. DMS also has lower measured inference latency than Marginal Sharpening in all twelve settings; timing details are provided in Appendix~\ref{app:time-cost}.

\subsection{Ablation Study on AIME24/25}

\label{sec:component-analysis}

We examine the contributions of the two sharpening stages and island sampling on AIME24/25 using Qwen3-8B. All two-stage variants share the same prompts, grader, and 16,384-token limit per reasoning--answer trajectory, with $K=16$ reasoning particles and $M=32$ answer candidates. In \Cref{tab:aime-component-analysis}, the reference DMS configuration uses $(\alpha_Z,\alpha_A)=(1.005,4)$. Setting either exponent to one disables the corresponding sharpening operation; setting both to one gives structured decoding without sharpening. Island-DMS keeps the exponents and total particle count fixed while dividing the reasoning population into eight independent islands of two particles.

\begin{table}[h]
\centering
\small
\caption{Component analysis on AIME24/25 using Qwen3-8B. Expected accuracy is the probability of selecting a correct answer under the normalized answer weights, averaged over problems. Pass@1 evaluates one sampled answer per problem, while oracle pass@32 is the fraction of problems with at least one correct answer among the 32 candidates. All accuracy metrics are percentages.}
\label{tab:aime-component-analysis}
\begin{tabular}{lccccc}
\toprule
Variant
& $\alpha_Z$
& $\alpha_A$
& Expected
& Pass@1
& Oracle pass@32 \\
\midrule

Structured decoding only
& 1 & 1 & 66.48 & 67.78 & \textbf{82.78} \\

Answer sharpening only
& 1 & 4 & \textbf{76.89} & \textbf{77.22} & \textbf{82.78} \\

Reasoning sharpening only
& 1.005 & 1 & 71.77 & 72.78 & 78.33 \\

Standard DMS
& 1.005 & 4 & 74.55 & 74.44 & 78.33 \\

Island-DMS
& 1.005 & 4 & 76.83 & \textbf{77.22} & 81.67 \\
\bottomrule
\end{tabular}
\end{table}

With unsharpened reasoning ($\alpha_Z=1$), increasing $\alpha_A$ from 1 to 4 raises expected accuracy from 66.48\% to 76.89\%. At $\alpha_A=1$, mild reasoning sharpening raises average candidate correctness from 66.48\% to 71.77\%, but reduces oracle pass@32 from 82.78\% to 78.33\%. Reasoning sharpening therefore improves average candidate correctness while reducing correct-answer coverage in this setting. At fixed exponents and particle count, Island-DMS raises expected accuracy from 74.55\% to 76.83\% and oracle pass@32 from 78.33\% to 81.67\%. However, answer sharpening alone achieves similar expected accuracy (76.89\%), and both obtain 77.22\% pass@1. Thus, additional reasoning sharpening is not required for the best observed pass@1 in this setting, supporting independent choices of the two exponents.

Due to space limit, we defer additional experiments to the appendix. In particular, we report sensitivity to the two sharpening exponents in Appendix~\ref{app:alpha-sensitivity} and ablations of annealing and resampling in Appendix~\ref{app:annealing-resampling}. We also provide a diagnostic using a higher-budget reasoning sampler in Appendix~\ref{app:high-fidelity-diagnostic}, together with inference latency comparison and a breakdown of DMS runtime in Appendix~\ref{app:time-cost}.

\section{Conclusion}

We studied training-free inference-time scaling for LLM reasoning, aiming to improve answer accuracy without updating model weights or relying on external reward models. Sequence-level power sampling faces both sampler-level and target-level challenges: finite-particle SMC can suffer from path collapse, while full-sequence sharpening can suppress alternative reasoning paths that support the same answer. We proposed \textbf{DMS}, a two-stage marginal sharpening framework with separate sampler-level and target-level designs. At the sampler level, FAA-SMC combines annealed power targets, predictive resampling, and locally fully adapted proposals. At the target level, DMS sharpens the reasoning distribution, marginalizes answer probabilities over the resulting reasoning population, and then sharpens the induced answer marginal. Separate exponents control reasoning-path concentration and answer-level consensus, allowing answers to be weighted according to their aggregate support across reasoning paths. Experiments on MATH500, GPQA-Diamond, HumanEval, and AIME24/25 across three base models show that DMS achieves the highest mean pass@1 among the compared methods in all twelve settings, including comparisons with Power-SMC and Marginal Sharpening. Future work includes improving answer-marginal estimation, reducing particle degeneracy during reasoning, and developing adaptive choices of sharpening strengths and inference-time budgets.

\appendix

\section{Additional Details for DMS}
\label{app:dms-details}

Algorithm~\ref{alg:faa_smc} summarizes FAA-SMC. DMS applies this sampler to the reasoning segment with terminal exponent $\alpha=\alpha_Z$.

\paragraph{Target Preservation under Annealing.}
Let $Q(y_{1:T}\mid x)=\prod_{t=1}^Tq_t(y_t\mid x,y_{<t})$ denote the sequence proposal obtained from the token-level proposal kernels. Along any fixed sequence with positive proposal probability, the incremental importance ratios telescope:
\begin{align*}
    \prod_{t=1}^T G_t(y_{1:t}) = \frac{\gamma_T(y_{1:T}\mid x)}{\gamma_0(\varnothing\mid x)Q(y_{1:T}\mid x)} = \frac{p_0(y_{1:T}\mid x)^\alpha}{Q(y_{1:T}\mid x)}.
\end{align*}
Thus, annealing changes the intermediate targets while preserving the specified final power target. For a terminated particle, subsequent steps append the fixed terminal symbol with probability one, so $\mathcal{Z}_t(h_t)=1$. Its incremental corrections then reduce to $p_0(y_{<t}\mid x)^{\beta_t-\beta_{t-1}}$, which must still be applied until the final exponent is reached.

\begin{algorithm}[h]
\caption{FAA-SMC: Fully Adapted Annealed SMC}
\label{alg:faa_smc}
\begin{algorithmic}[1]
\REQUIRE Prompt $x$, base model $p_0$, exponent schedule $1=\beta_0\leq\cdots\leq\beta_T=\alpha$, particle count $K$, resampling threshold $\tau$, and resampling checkpoints.
\STATE Initialize $Y_{1:0}^k=\varnothing$ and $w_0^k=1$ for $k=1,\dots,K$.
\FOR{$t=1,\dots,T$}
    \STATE Set $h_t^k=(x,Y_{<t}^k)$ and compute $q_t^{\mathrm{FA}}(\cdot\mid h_t^k)$ and $M_t(h_t^k)$ for all $k$.
    \STATE Compute $\widetilde w_{t-1}^k=w_{t-1}^kM_t(h_t^k)$ and normalize to obtain $\widetilde{\bar w}_{t-1}^{1:K}$.
    \IF{$t$ is a resampling checkpoint and $\operatorname{ESS}(\widetilde{\bar w}_{t-1}^{1:K})<\tau K$}
        \STATE Resample the prefixes and their associated proposals according to $\widetilde{\bar w}_{t-1}^{1:K}$, and set $w_t^k=1$ for all $k$.
    \ELSE
        \STATE Set $w_t^k=\widetilde w_{t-1}^k$ for all $k$.
    \ENDIF
    \STATE For prefixes that terminated before step $t$, set $Y_t^k$ to the fixed terminal symbol.
    \STATE For all other particles, independently sample $Y_t^k\sim q_t^{\mathrm{FA}}(\cdot\mid x,Y_{<t}^k)$.
\ENDFOR
\STATE Normalize the final weights and \textbf{return} $\{(Y_{1:T}^k,\bar w_T^k)\}_{k=1}^K$.
\end{algorithmic}
\end{algorithm}

\paragraph{Augmented Representation of the Answer Target.}
The answer target in \eqref{equ:tms_answer_target} admits the joint representation
\begin{align*}
    \widetilde\pi_A(z,a\mid x) = \mathcal{Z}_A(x)^{-1}\pi_Z(z\mid x)p_0(a\mid z,x)m_{\alpha_Z}(a\mid x)^{\alpha_A-1}.
\end{align*}
Summing over reasoning paths gives
\begin{align*}
    \sum_z \widetilde\pi_A(z,a\mid x) = \mathcal{Z}_A(x)^{-1}m_{\alpha_Z}(a\mid x)^{\alpha_A} = \pi_A(a\mid x).
\end{align*}
The additional power factor depends on the answer marginal and is shared by all reasoning paths paired with the same answer sequence.

\paragraph{Numerical Evaluation of Answer Weights.}
We compute the answer importance weights using the empirical marginal $\widehat m_K$, which approximates $m_{\alpha_Z}$ from the finite weighted reasoning population. For each candidate $A_j$, its conditional log probability under reasoning path $z_i$ is
\begin{align*}
    \ell_{j,i} = \log p_0(A_j\mid z_i,x) = \sum_{t=1}^{|A_j|}\log p_0(A_{j,t}\mid x,z_i,A_{j,<t}).
\end{align*}
Only answer tokens contribute to this score; the reasoning path provides the conditioning context. To evaluate the empirical marginal in log space, we set $b_j=\max_{1\leq i\leq K}(\log\bar w_i^Z+\ell_{j,i})$ and compute
\begin{align*}
    \log\widehat m_K(A_j\mid x) = b_j+\log\left(\sum_{i=1}^K \exp\left(\log\bar w_i^Z+\ell_{j,i}-b_j\right)\right).
\end{align*}
The log importance weight is $\log w_j^A=(\alpha_A-1)\log\widehat m_K(A_j\mid x)$. We similarly normalize these weights using $c=\max_{1\leq r\leq M}\log w_r^A$:
\begin{align*}
    \bar w_j^A = \frac{\exp(\log w_j^A-c)}{\sum_{r=1}^M\exp(\log w_r^A-c)}.
\end{align*}
Subtracting the maximum improves numerical stability without changing the marginal probabilities or normalized importance weights.

\section{Experimental Setup and Implementation Details}
\label{app:experimental-details}

\paragraph{Models and benchmarks.}
We evaluate Qwen3-4B, Qwen3-8B, and DeepSeek-R1-Distill-Llama-8B using their native chat templates, with thinking mode enabled for Qwen3. Inference uses \texttt{bfloat16} on one NVIDIA H100 80\,GB GPU per task. We evaluate all 500 MATH500 problems, 198 GPQA-Diamond questions, 164 HumanEval problems, and 60 AIME24/25 problems. Results are averaged over seeds $\{10080,10081,10082\}$.

\paragraph{Generation and evaluation.}
Within each model--benchmark setting, all methods use the same model checkpoint, tokenizer, user-level problem prompt, maximum length per reasoning--answer trajectory, answer extractor, and grader. The maximum generation lengths are $4224$ tokens for MATH500, $8192$ for GPQA-Diamond, $4608$ for HumanEval, and $16384$ for AIME24/25. Standard and low-temperature decoding generate a single completion containing both reasoning and the final answer, using the full generation limit. Power-SMC likewise generates complete reasoning--answer particles within the same per-particle limit. These methods do not reserve a separate answer budget or insert a reasoning--answer boundary. Marginal Sharpening and DMS allocate the trajectory budget as $4096+128$ tokens for MATH500, $8064+128$ for GPQA-Diamond, $4096+512$ for HumanEval, and $16256+128$ for AIME24/25, where the first and second terms specify the reasoning and answer limits, respectively. If a reasoning trace reaches its limit without producing a reasoning--answer boundary, we append a single \texttt{</think>} token. This token is included in the continuation context but excluded from the reasoning-path likelihood and residual importance-weight calculation. For non-code benchmarks, the separated answer stage is initialized with a boxed-answer prefix and terminates after a complete boxed expression; HumanEval uses a native code continuation. Answer generation and cross-scoring use the same continuation context, including any inserted boundary token and fixed answer prefix; only generated answer tokens contribute to the conditional answer likelihood. Mathematical predictions are evaluated by normalized exact matching and symbolic equivalence, GPQA uses exact option matching, and HumanEval executes the original base tests in an isolated process with a three-second timeout.

\paragraph{Baselines.}
Standard and low-temperature decoding use temperatures $1$ and $0.25$, respectively, with top-$p=1$. Power-SMC uses $N=16$ particles and checks the particle weights every 64 tokens, applying systematic resampling when $\operatorname{ESS}<0.5N$. Its final power exponents for (MATH500, GPQA-Diamond, HumanEval, AIME24/25) are $(4,1.05,4,4)$ for Qwen3-4B, $(2,1.25,4,4)$ for Qwen3-8B, and $(1.5,1.01,1.01,1.01)$ for DeepSeek-R1-Distill-Llama-8B. By default, the proposal uses temperature $0.25$ and top-$p=0.9$, and the exponent increases linearly from $1$ to its final value over 100 tokens. For Qwen3-8B on GPQA-Diamond and DeepSeek-R1-Distill-Llama-8B on GPQA-Diamond, HumanEval, and AIME24/25, the proposal instead uses temperature $1$ and top-$p=1$, with a 256-token linear annealing schedule. Marginal Sharpening uses a sharpening exponent of $4$ and $8$ independent Monte Carlo groups, each containing $4$ reasoning traces, for $32$ traces in total. It generates a single answer token by token using the resulting prefix-weighted marginal token probabilities.

\paragraph{DMS configuration.}
Unless stated otherwise, DMS uses $K=16$ reasoning particles and $M=32$ answer candidates. The power-two annealing schedule begins after 1024 reasoning tokens and reaches the terminal exponent at the benchmark-specific maximum reasoning length. For standard DMS, we check the ESS of the predictive weights every 64 tokens and apply systematic resampling when it falls below $K/2$. Each complete answer candidate is cross-scored under all retained reasoning contexts, and the final answer is sampled according to its normalized answer importance weight. Table~\ref{tab:dms-hyperparameters} lists the exponents used in the main results.

\begin{table}[h]
\centering
\small
\setlength{\tabcolsep}{5pt}
\caption{DMS reasoning and answer exponents $(\alpha_Z,\alpha_A)$ used in the main results (\Cref{tab:main-results}).}
\label{tab:dms-hyperparameters}
\begin{tabular}{lcccc}
\toprule
Model & MATH500 & GPQA-Diamond & HumanEval & AIME24/25 \\
\midrule
Qwen3-4B & $(1.25,3)$ & $(1.025,2)$ & $(1.25,1.5)$ & $(1.025,3.5)$ \\
Qwen3-8B & $(1.025,2.5)$ & $(1.025,2)$ & $(1.25,1.5)$ & $(1.005,4)$ \\
DeepSeek-R1-Distill-Llama-8B & $(1.001,6)$ & $(1.001,5)$ & $(1.25,1.5)$ & $(1.001,8)$ \\
\bottomrule
\end{tabular}
\end{table}

\paragraph{Island implementation.}
On AIME24/25, all three models use eight islands of two particles, giving $K=16$ reasoning particles in total. Resampling occurs independently within each island. 
At the reasoning--answer boundary, we normalize the particle weights within each island and assign equal total weight to all islands. Specifically, particle $k$ in island $\ell$ receives pooled weight $\bar w_{\ell k}/L$, where $\sum_k\bar w_{\ell k}=1$ and $L=8$. These pooled weights define the empirical answer marginal used for candidate generation and selection.

\section{Sensitivity to the Two Sharpening Exponents}
\label{app:alpha-sensitivity}

We study the interaction between reasoning and answer sharpening using standard DMS with Qwen3-8B on AIME24/25. We evaluate
\begin{align*}
    \alpha_Z \in \{1,1.001,1.005,1.01,1.025\}, \qquad \alpha_A \in \{1,1.5,2,2.5,3,3.5,4,5,6,8\}.
\end{align*}
All configurations use $K=16$, $M=32$, and the same generation budgets and sampler settings. 

We report expected accuracy, the probability that categorical selection returns a correct answer, averaged over problems. For candidate correctness indicators $c_m(x)\in\{0,1\}$ and normalized answer weights $\bar w_m(x;\alpha_A)$, this is
\begin{align*}
    \operatorname{Acc}_{\mathrm{exp}} = \frac{1}{|\mathcal D|}\sum_{x\in\mathcal D}\sum_{m=1}^{M}\bar w_m(x;\alpha_A)c_m(x),
\end{align*}
where $\mathcal D$ is the benchmark problem set. We average this metric over three seeds. At $\alpha_A=1$, all answer weights are equal, so expected accuracy equals average candidate correctness. Oracle pass@32 is the fraction of problems with at least one correct candidate and remains unchanged across answer exponents within a fixed bank.

\begin{figure}[h]
    \centering
    \includegraphics[width=\linewidth]{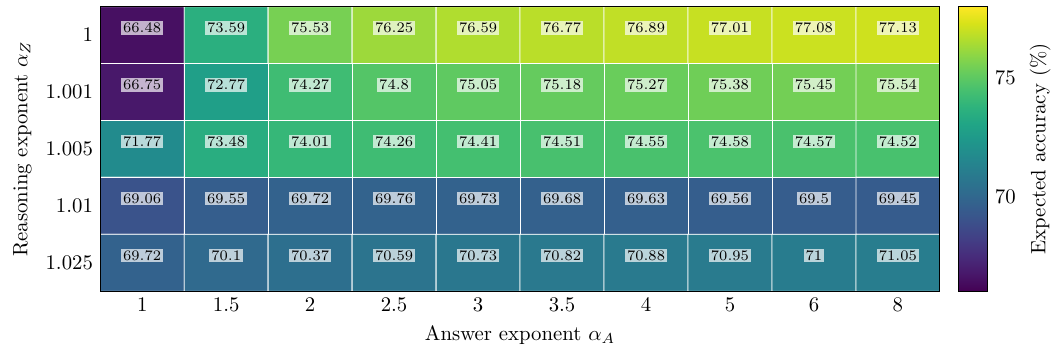}
    \caption{Expected accuracy (\%) of standard DMS with Qwen3-8B on AIME24/25 across reasoning and answer exponents, averaged over three seeds.}
    \label{fig:aime-alpha-sensitivity}
\end{figure}

Figure~\ref{fig:aime-alpha-sensitivity} shows that answer sharpening improves expected accuracy over uniform answer selection across all tested reasoning exponents. At $\alpha_Z=1$, expected accuracy rises from $66.48\%$ at $\alpha_A=1$ to $76.89\%$ at $\alpha_A=4$ and $77.13\%$ at $\alpha_A=8$. Thus, most of the improvement is obtained by $\alpha_A=4$, with limited additional gains from stronger answer sharpening.

Mild reasoning sharpening can also improve average candidate correctness: setting $\alpha_Z=1.005$ raises it from $66.48\%$ to $71.77\%$. However, the highest expected accuracy obtained from this reasoning bank is $74.58\%$, below the $77.13\%$ achieved with $\alpha_Z=1$. Improving average candidate correctness therefore does not necessarily improve accuracy after answer reweighting. In this standard (non-island) setting, the strongest result combines an unsharpened reasoning distribution with a larger answer exponent. These results support choosing the two sharpening strengths independently, allowing strong answer sharpening without requiring additional concentration of the reasoning distribution.

\section{Annealing and Resampling Ablations}
\label{app:annealing-resampling}

We evaluate annealing and reasoning-stage resampling in standard DMS using Qwen3-8B on AIME24/25, with $K=16$, $M=32$, and $(\alpha_Z,\alpha_A)=(1.005,4)$. The reference sampler uses a quadratic annealing schedule that remains at $\beta_t=1$ through token 1024 and reaches $\alpha_Z$ at the maximum reasoning length. At 64-token checkpoints, it performs systematic resampling whenever the predictive ESS falls below $K/2$.

The no-annealing variant applies $\alpha_Z$ from the first generated token, using the locally fully adapted proposal $q_t^{\mathrm{FA}}(v\mid h_t)\propto p_0(v\mid h_t)^{\alpha_Z}$ and the corresponding incremental importance weights. The no-resampling variant retains the reference annealing schedule, proposal, and sequential weight updates, but disables ancestor resampling. Its final reasoning weights are still used to construct the empirical answer marginal.

All variants approximate the same terminal reasoning target. Completed reasoning traces stop generating tokens but continue to receive annealing weight corrections as $\beta_t$ increases. If all particles finish before the terminal exponent is reached, the remaining annealing updates are completed without generating additional tokens.

\begin{table}[h]
\centering
\small
\setlength{\tabcolsep}{4pt}
\caption{Sampler ablations for non-island DMS with Qwen3-8B on AIME24/25, averaged over three seeds. Expected accuracy and oracle pass@32 are percentages. Final ESS is computed from the final reasoning weights and divided by $K=16$. Resamples and unique roots report the mean number of reasoning-stage resampling events and surviving initial-particle ancestors, respectively.}
\label{tab:aime-annealing-ablation}
\begin{tabular}{lccccc}
\toprule
Variant & Expected & Oracle pass@32 & Final ESS$/K$ & Resamples & Unique roots \\
\midrule
standard DMS & \textbf{74.55} & 78.33 & 0.765 & 1.39 & 8.17 \\
No annealing & 71.72 & 77.78 & 0.773 & 2.26 & 7.11 \\
No resampling & 72.09 & \textbf{80.00} & 0.252 & 0 & \textbf{16.00} \\
\bottomrule
\end{tabular}
\end{table}

Table~\ref{tab:aime-annealing-ablation} shows that the reference sampler achieves the highest mean expected accuracy, exceeding the no-annealing and no-resampling variants by 2.83 and 2.46 percentage points, respectively. Without annealing, resampling occurs more frequently and fewer distinct initial ancestors remain in the final population. This pattern is consistent with gradual annealing limiting ancestor loss during resampling.

Disabling resampling preserves all 16 initial ancestors and increases oracle pass@32 from $78.33\%$ to $80.00\%$. However, the final reasoning weights become more concentrated, with normalized ESS falling to $0.252$, and expected accuracy decreases to $72.09\%$. Thus, preserving more ancestors and greater correct-answer coverage does not necessarily improve weighted answer selection.

Final ESS and the number of surviving ancestors measure different aspects of the reasoning population. Resampling can balance particle weights without recovering ancestors that have already been discarded. Similar final ESS values can therefore coexist with different numbers of surviving ancestors, as observed for the reference and no-annealing variants.

\section{Higher-Budget Reasoning-Sampler Diagnostic}
\label{app:high-fidelity-diagnostic}

We further examine sensitivity to reasoning sharpening using a higher-budget replica-exchange Markov chain Monte Carlo sampler in place of FAA-SMC. Each replica targets $\pi_{\alpha_Z}(z\mid x)\propto p_0(z\mid x)^{\alpha_Z}$ through local Metropolis--Hastings updates that revise previously generated reasoning traces. Adjacent replicas also propose exchanging their traces to improve mixing across the exponent ladder. We use the resulting finite-chain samples as a diagnostic reference rather than as exact samples from the target distributions.

We evaluate Qwen3-4B on HumanEval using seeds $\{10080,10081,10082\}$, with reasoning and answer limits of 4096 and 512 tokens, respectively. For each problem and seed, we run 16 independent replica ladders and retain one final reasoning trace per exponent from each ladder. After the initial generation block, each replica receives 50 local Metropolis--Hastings updates. Each update selects a restart position uniformly within the generated reasoning trace and regenerates the remaining suffix at temperature $1/\alpha_Z$, followed by the corresponding Metropolis--Hastings acceptance step. After each round of local updates, the sampler performs a complete sweep of adjacent replica-exchange proposals. We run the fine ladder $(1.1,1.05,1.025,1)$ and the extended ladder $(4,3,2,1.5,1.25,1.1)$ independently. These configurations maintain 64 and 96 replica chains per problem and seed, respectively, in addition to the repeated suffix generation required by the local updates.

\begin{table}[h]
\centering
\small
\caption{Reasoning-exponent sensitivity with a higher-budget replica-exchange sampler using Qwen3-4B on HumanEval. Expected accuracy and oracle pass@32 are reported as percentages averaged over three seeds. The fine and extended ladders are run independently.}
\label{tab:high-fidelity-diagnostic}
\begin{tabular}{ccc}
\toprule
$\alpha_Z$ & Expected accuracy ($\alpha_A=4$) & Oracle pass@32 \\
\midrule
\multicolumn{3}{l}{Fine ladder} \\
$1$ & $95.06$ & $98.17$ \\
$1.025$ & $96.17$ & $97.97$ \\
$1.05$ & $96.59$ & $97.97$ \\
$1.1$ & $96.39$ & $97.97$ \\
\midrule
\multicolumn{3}{l}{Extended ladder} \\
$1.1$ & $96.06$ & $98.17$ \\
$1.25$ & $96.02$ & $98.17$ \\
$1.5$ & $96.61$ & $98.17$ \\
$2$ & $95.88$ & $97.97$ \\
$3$ & $96.01$ & $97.97$ \\
$4$ & $96.23$ & $97.97$ \\
\bottomrule
\end{tabular}
\end{table}

For each reasoning exponent, the $K=16$ retained traces define a uniformly weighted reasoning mixture. We generate $M=32$ answer candidates from this mixture and cross-score each candidate under all retained reasoning contexts. The same candidates and cross-score matrix are reused across the answer-exponent grid. Candidates are evaluated using the same full-program-preserving HumanEval extractor and isolated execution protocol as the main results, with a three-second timeout. Table~\ref{tab:high-fidelity-diagnostic} reports expected accuracy at $\alpha_A=4$ and oracle pass@32.

Both ladders achieve their highest expected accuracy at moderate reasoning exponents within their respective ranges. In the fine ladder, expected accuracy increases from $95.06\%$ at $\alpha_Z=1$ to $96.59\%$ at $\alpha_Z=1.05$, before slightly decreasing to $96.39\%$ at $\alpha_Z=1.1$. In the extended ladder, expected accuracy likewise peaks at a moderate exponent, reaching $96.61\%$ at $\alpha_Z=1.5$.

Oracle pass@32 decreases slightly from $98.17\%$ to $97.97\%$ as $\alpha_Z$ increases within each ladder. This pattern is consistent with the trade-off introduced by sharpening: concentrating probability on more likely reasoning paths can reduce diversity and the chance of finding a `lucky' correct answer through a less likely path. Meanwhile, the gains in expected accuracy show that moderate reasoning sharpening can improve weighted answer selection while maintaining high correct-answer coverage.

\section{Time-Cost Analysis}
\label{app:time-cost}

DMS inference consists of reasoning-particle generation, answer-candidate generation, and cross-scoring the $K\times M$ reasoning--answer pairs. Cross-scoring reuses cached reasoning prefixes to evaluate answer continuations. We measure wall-clock inference time per problem using the hardware configuration described in Appendix~\ref{app:experimental-details}. The measurements include all three stages and exclude model loading and offline grading.

\begin{table*}[h]
\centering
\small
\setlength{\tabcolsep}{7pt}
\renewcommand{\arraystretch}{1.12}
\caption{Mean wall-clock inference time in seconds per problem, averaged over seeds.}
\label{tab:main-results-runtime}
\resizebox{\textwidth}{!}{
\begin{tabular}{llrrrr}
\toprule
\textbf{Model} & \textbf{Method} & \textbf{MATH500}
& \textbf{GPQA-Diamond} & \textbf{HumanEval} & \textbf{AIME24/25} \\
\midrule
\multirow{5}{*}{\textbf{Qwen3-4B}}
& Standard Decoding        & 50.57 & 124.41 & 43.06 & 249.27 \\
& Low-Temperature Decoding & 49.36 & 123.26 & 43.09 & 259.47 \\
& Power-SMC                & 46.01 & 163.14 & 43.33 & 376.40 \\
& Marginal Sharpening      & 140.51 & 710.42 & 103.05 & 1957.74 \\
& \textbf{DMS (Ours)}
& 62.09 & 241.09 & 83.49 & 617.09 \\
\midrule
\multirow{5}{*}{\textbf{Qwen3-8B}}
& Standard Decoding        & 57.09 & 133.99 & 43.92 & 261.37 \\
& Low-Temperature Decoding & 52.62 & 131.98 & 44.05 & 269.33 \\
& Power-SMC                & 52.36 & 191.54 & 44.95 & 369.65 \\
& Marginal Sharpening      & 477.17 & 1270.15 & 436.42 & 2855.41 \\
& \textbf{DMS (Ours)}
& 82.69 & 237.56 & 89.30 & 874.87 \\
\midrule
\multirow{5}{*}{\textbf{DeepSeek-R1-Distill-Llama-8B}}
& Standard Decoding        & 33.13 & 92.97 & 34.79 & 172.13 \\
& Low-Temperature Decoding & 28.47 & 93.20 & 32.66 & 177.10 \\
& Power-SMC                & 31.97 & 177.11 & 54.48 & 483.29 \\
& Marginal Sharpening      & 359.23 & 1002.83 & 389.05 & 2204.19 \\
& \textbf{DMS (Ours)}
& 76.03 & 207.87 & 132.94 & 740.26 \\
\bottomrule
\end{tabular}
}
\end{table*}

Table~\ref{tab:main-results-runtime} reports the mean inference time per problem. Relative to standard decoding, non-island DMS requires $1.23$--$3.82$ times the latency, while Island-DMS on AIME24/25 requires $2.48$--$4.30$ times the latency. Across all twelve model--benchmark settings, DMS has lower measured latency than Marginal Sharpening.

For Island-DMS with Qwen3-8B on AIME24/25, reasoning generation, answer generation, and cross-scoring take $705.38$\,s ($80.63\%$), $25.24$\,s ($2.89\%$), and $144.23$\,s ($16.49\%$) per problem, respectively. Reasoning generation accounts for more than four-fifths of the measured latency, followed by cross-scoring, while answer generation contributes less than $3\%$. These timings reflect the evaluated implementations and configurations.

\bibliography{reference}
\bibliographystyle{ims}

\end{document}